\documentclass{article}

\usepackage[final]{neurips_2024}
\usepackage[utf8]{inputenc} 
\usepackage[T1]{fontenc}    
\usepackage{hyperref}       
\usepackage{url}            
\usepackage{booktabs}       
\usepackage{amsfonts}       
\usepackage{nicefrac}       
\usepackage{microtype}      
\usepackage{xcolor}         
\usepackage[pdftex]{graphicx}
\usepackage{amsmath}
\usepackage{algorithm}
\usepackage{algpseudocode}
\usepackage{sidecap}
\usepackage{MnSymbol}

\newcommand{\argmaxtopk}{\mathop{\mathrm{arg\!max}_\mathrm{topk}}}

\title{Harnessing Preference Optimisation in Protein LMs for Hit Maturation in Cell Therapy}

\author{%
Katarzyna Janocha \quad Annabel Ling \quad Alice Godson \quad Yulia Lampi \\
\textbf{Simon Bornschein} \quad \textbf{Nils Y.\ Hammerla} \\
Coding Bio \\
\texttt{\{kasia,annabel,alice,yulia,simon,nils\}@coding.bio}\\
}

\begin{document}

\maketitle

\begin{abstract}
Cell and immunotherapy offer transformative potential for treating diseases like cancer and autoimmune disorders by modulating the immune system. The development of these therapies is resource-intensive, with the majority of drug candidates failing to progress beyond laboratory testing. While recent advances in machine learning have revolutionised areas such as protein engineering, applications in immunotherapy remain limited due to the scarcity of large-scale, standardised datasets and the complexity of cellular systems. In this work, we address these challenges by leveraging a high-throughput experimental platform to generate data suitable for fine-tuning protein language models. We demonstrate how models fine-tuned using a preference task show surprising correlations to biological assays, and how they can be leveraged for few-shot hit maturation in CARs. This proof-of-concept presents a novel pathway for applying ML to immunotherapy and could generalise to other therapeutic modalities.
\end{abstract}

\section{Introduction}
Cell and immunotherapy hold tremendous promise for treating traditionally incurable diseases, with applications ranging from oncology and autoimmune disorders to age-related conditions. The overarching goal is to modulate the patient’s immune response, either by directly engineering immune cells or by introducing engineered proteins to enhance or suppress immune function. The main modality we will explore in this work are Chimeric Antigen Receptors (CARs) - engineered proteins that are expressed on the surface of T-cells. CARs act as artificial receptors designed to bind to a specific antigen (target), for instance on the cell surface of a tumor, and trigger an immune response.

The development of immunotherapies, and drug development in general, is a resource-intensive process. Despite extensive efforts to evaluate candidates in the lab, the great majority of identified drug compounds will fail to progress through clinical trials to become viable treatments for patients \citep{mullard2016parsing}. To improve the performance of candidate drug compounds, the field of drug discovery has increasingly turned to computational approaches to make better use of experimental data and facilitate exploration of the vast design space for such molecules.

Over recent years, the modelling of discrete time-series such as natural text has seen tremendous progress, where the capabilities of transformer-based language models \citep{vaswani2017attention} far exceeded the expectations by domain experts. Naturally, these methods have been applied to protein engineering, aiming to construct biologically relevant representations of amino-acid (or tokenised DNA-) sequences via masking \citep{lin2023evolutionary, hayes2024simulating} or auto-regressive Protein Language Models (PLMs) \citep{nijkamp2023progen2}. Areas such as structure prediction \cite{abramson2024accurate}, structure generation \citep{watson2023novo}, and protein-protein interactions such as docking \citep{corso2022diffdock} have seen significant progress via large and sophisticated ML models based on transformers or diffusion generative models, among other approaches. This progress is largely spurred by the availability of relevant data at scale \citep{suzek2015uniref,steinegger2018clustering} and clearly defined research questions, which has led to a rich and mature ecosystem of academic and commercial interest around ML applications in protein engineering.

Another area of focus has been around in-silico design and optimisation of antibodies \citep{joubbi2024antibody}, with large datasets available for academic and commercial use \ \citep{olsen2022observed}. Applications include the virtual screening of potential binders \citep{bachas2022antibody}, affinity maturation \citep{ruffolo2021deciphering,clark2023enhancing, hie2024efficient} or the optimisation of enzymes \citep{he2024protein}. Much of the work in this space is driven by the availability of language models trained on antibody data, including masking \citep{ruffolo2021deciphering, Olsen2022, prihoda2022biophi} as well as auto-regressive models \citep{shuai2021generative,nijkamp2023progen2}.

In contrast, ML applications in immunotherapy are relatively scarce, and have predominantly relied on basic ML such as motif-based optimisation of CARs \citep{castellanos2022speedingcars,daniels2022decoding}, with more sophisticated modelling only being explored recently for bispecific antibodies \citep{mullin2024applications}. This is partly due to the lack of publicly accessible, standardized datasets, as well as the inherent complexity of the problem compared to small molecule drug design. The data in this field are influenced by a multitude of factors related to living cells, making it challenging to generate datasets at a scale sufficient for effective training or fine-tuning of large models. The field thus continues to rely on manual, predominantly low-throughput approaches to drug discovery.

In this work we attempt to bridge this gap, demonstrating how large-scale, low-precision yet high throughput testing of thousands of drug compounds in cells can be harnessed to generate data suitable for fine-tuning PLMs for the task of hit maturation in CARs. We show, perhaps surprisingly, that model loss of an auto-regressive PLM, fine-tuned using a preference task, is highly correlated to the results of standard biological assays, and that we can reliably find improved mutants via few-shot exploration of the design space. This provides a proof-of-concept that such models can effectively guide the design of CARs, which could generalise to other modalities in immunotherapy.

\section{Chimeric Antigen Receptors (CARs)}
\begin{figure}
  \includegraphics[width=0.7\linewidth]{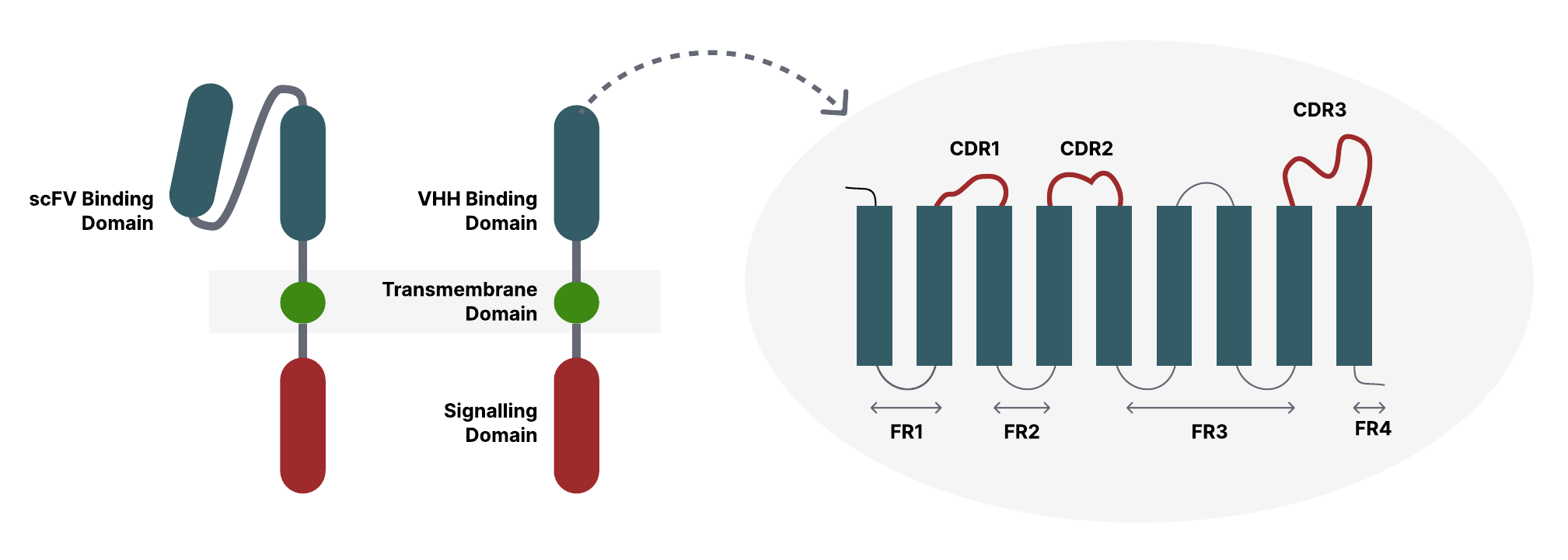}
  \centering
  \caption{CAR structure with scFV and VHH binding domains (see text for details).}
  \label{fig:car}
\end{figure}

Briefly, immunotherapies use a variety of mechanisms to harness the immune system to recognise and eliminate diseases. These immunotherapies include monoclonal antibodies (mAb), checkpoint inhibitors, T cell engagers and CAR-T cell therapy. CAR-T cell therapy has predominantly been used to treat haematological malignancies such as multiple myeloma and leukemia, where it may lead to complete remission in a significant fraction of patients \citep{cappell2023long}. It also shows potential in other malignant diseases and autoimmune diseases \citep{blache2023car}.

Typically, a CAR consists of a binding domain which binds to antigens on tumour cells, a hinge domain that gives the binder flexibility, a transmembrane domain that anchors the CAR to the T cell membrane, and a signalling domain which  facilitates T cell activation (Figure \ref{fig:car}). 

Various types of binding domains can be presented on the CAR, but most common are those based on single-chain fragment variable (scFv) of mAb. In contrast, here we explore the use and optimisation of variable heavy domain of heavy chain (VHH) fragments, which are antibody fragments naturally derived from the Camelidae family. scFvs and VHH fragments are comparable  in terms of affinity \citep{muyldermans2013nanobodies} yet the reduced size of VHHs promotes advantageous stability, solubility and immunogenicity characteristics \citep{gorovits2019immunogenicity}.

VHHs consist of three hypervariable complementary determining regions (CDRs) interspersed with framework regions (FRs), as depicted in Figure \ref{fig:car}. CDRs play a critical role in antigen binding, with their flexible structures adapting to fit the binding site. In contrast, FRs serve as a scaffold to support the CDRs for effective antigen binding, and therefore exhibit significantly less sequence diversity in comparison to CDRs. CDR3 is the most diverse region and has a longer length distribution in VHHs than in scFvs, allowing for more versatility in targeting antigen binding sites.

For CAR T-cell therapy to be efficacious, the designed protein must fulfil several criteria. The protein must be successfully transduced into T-cells via a suitable vector, express well, bind specifically to the target antigen, and finally trigger T-cell activation, ideally in a target-dependent manner. This complex behaviour makes the discovery or design of CARs a challenging problem.

\section{Hit maturation of CARs using ML}

In this work we explore Protein Language Models (PLMs) as a potential avenue to enable hit maturation in CARs - efficient, ML-guided exploration of the vast design space, going beyond naive approaches such as deep mutational scans \citep{fowler2014deep} or substitution methods \citep{henikoff1992amino}. We focus on the performance of CARs against a variety of targets, to discover patterns that either enable good performance (such as activation, expression, and specificity) or that negatively impact the cell.

We specifically use auto-regressive models in this work, as they i) allow us to assess the likelihood of a sequence under the model, ii) allow efficient sampling of the distribution learned by the model in order to explore mutations around an initial candidate\footnote{Recently, \cite{hayes2024simulating} have demonstrated how masking LMs can be fine-tuned using similar approaches and how issues with sampling can be circumvented, which we are keen to explore in future work.}, iii) have well-established methods and tools for fine-tuning after pre-training, and iv) may enable zero-shot approaches to the design of CARs.

\subsection{High-throughput evaluation of CARs}
\begin{figure}
  \includegraphics[width=0.75\linewidth]{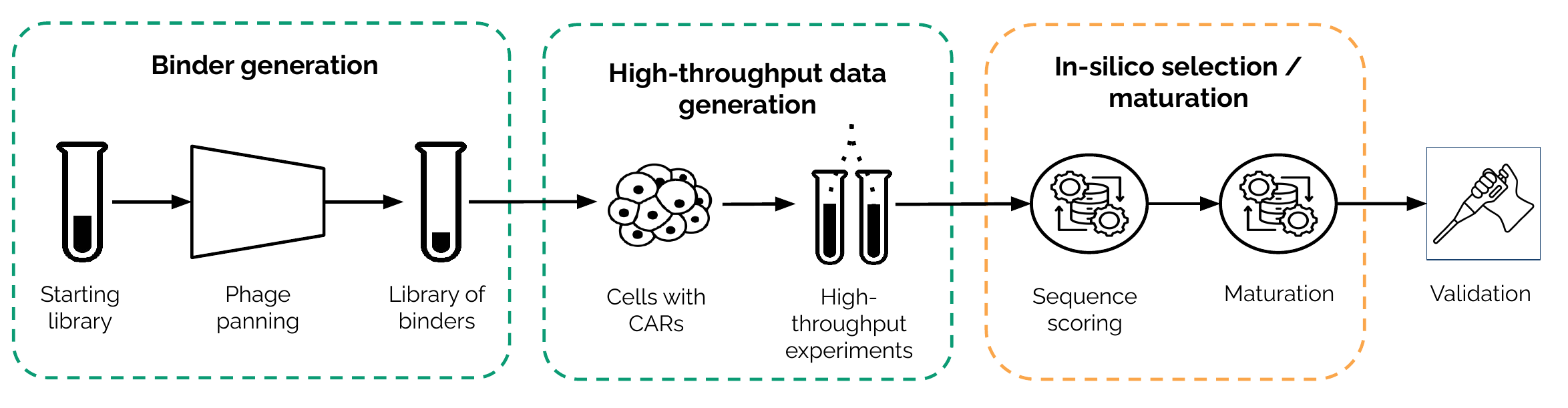}
  \centering
  \caption{Simplified candidate generation pipeline. Using phage display based on a highly diverse starting library we screen for binders to the target of interest using phage display. Resulting candidates are evaluated in cells in scalable proprietary assays.
  \vspace{-1em}}
  \label{fig:pipeline}
\end{figure}

To align PLMs to the task of hit maturation, we need to derive sufficient experimental data about the performance of CARs. The goal is to determine the suitability of a CAR for a specific target and to assign a scalar score that is well correlated with CAR performance.

Our platform, as depicted in Figure \ref{fig:pipeline}, starts of with a semi-synthetic VHH library, wherein most of the diversity ($\sim 10^{11}$) exists in the CDR3 region. We first isolate binders for a specific target by creating a VHH phage display library. These phages are exposed to immobilised target protein and non-specific binders are washed away  in a process known as phage panning. We conduct multiple rounds of this selection process to enrich the population of binders, and use Next Generation Sequencing (NGS) to identify the sequences of the binding population at each step.

The VHHs that bind to the target of interest are subsequently reformatted into a CAR library ($\sim10^{5}$). We then generate a pool of CAR-T cells, with each cell expressing a single CAR design. By co-culturing the library-expressing cells with target-expressing cells and target-negative cells, we can assess the level of cell activation and use a fluorescence-activated cell sorter to isolate the activated and non-activated subsets of these libraries. Using NGS we can identify the candidates in each subset. The fractions of activated and non-activated cells for each CAR across different cell lines are used to score their performance, leading to a single scalar associated with each CAR.

\subsection{Preference-based fine-tuning of PLMs}

In natural language, it is often challenging to assign absolute quality ratings to responses. Human raters can, however, express preferences between two outputs, establishing a partial order. \emph{Preference Optimization} has been developed to fine-tune large language models (LLMs) based on this signal, and techniques such as Reinforcement Learning with Human Feedback (RLHF) \citep{christiano2017deep} have become the most common approach for the alignment of LLMs. Our setting is similar -- high-throughput testing of CARs, while scalable, may lack sufficient precision to resolve the finest difference between candidate CARs. In line with other recent work in the field \citep{hayes2024simulating} we therefore formulate a preference-optimisation task.

We use \emph{Direct Preference Optimisation} (DPO) \citep{rafailov2024direct} to fine-tune models based on data derived from high-throughput experiments. DPO is a simplified approach to preference optimisation, which does not require a separate reward model, but instead formulates a simple classification problem to improve the probability that the chosen response is preferred over the rejected response. More formally, DPO relies on a sigmoid-based learning objective:

\begin{equation}
    \mathcal{L}_{\text{DPO}} (\pi_\theta \mid \pi_{\text{st}}, \mathcal{D}_p) = \mathbb{E}_{(x, y_w, y_u) \sim \mathcal{D}_p}  \left[ \log \sigma \left( \beta \log \frac{\pi_\theta (y_w \mid x)}{\pi_{\text{st}} (y_w \mid x)} - \beta \log \frac{\pi_\theta (y_u \mid x)}{\pi_{\text{st}} (y_u \mid x)} \right) \right]
\label{eq:dpo_loss}
\end{equation}
where $\pi_\text{st}$ is the pretrained model, $\pi_\theta$ the model being fine-tuned, $x$ is the context (prompt), and $y_w, y_u$ are the chosen and rejected completions.
The idea behind this loss is to maximise the margin between the rejected and chosen completions conditioned on the same context. We explore other loss functions beyond the sigmoid loss in equation \ref{eq:dpo_loss}:

\paragraph{Hinge,} as proposed by \citep{liu2024statistical} and originally inspired by \citep{zhao2023slic};
\begin{equation}
 \mathcal{L}_{\text{hinge}} (\pi_\theta \mid \pi_{\text{st}}, \mathcal{D}_p) = \mathbb{E}_{(x, y_w, y_u) \sim \mathcal{D}_p} \left[ \max \left( 0, 1 - \left( \beta \log \frac{\pi_\theta (y_w \mid x)}{\pi_{\text{st}} (y_w \mid x)} - \beta \log \frac{\pi_\theta (y_u \mid x)}{\pi_{\text{st}} (y_u \mid x)} \right) \right) \right]
\end{equation}

\paragraph{Kahneman-Tversky Optimization} \citep{ethayarajh2024kto}
\begin{equation}
\mathcal{L}_{\text{KTO}}(\pi_\theta \mid \pi_{\text{st}}, \mathcal{D}_p) = \mathbb{E}_{(x, y_w, y_u) \sim \mathcal{D}_p} \left[ 1 - v(x, y) \right]
\label{eq:kto_loss}
\end{equation}

where
$$ 
v(x, y) =
\begin{cases}
\sigma \left( \beta \left( \log \frac{\pi_\theta(y \mid x)}{\pi_{st}(y \mid x)} - \mathrm{KL}(\pi_\theta(y \mid x) \| \pi_{st}(y \mid x)) \right) \right) & \text{if } y = y_{\text{w}} \\
\sigma \left( \beta \left( \mathrm{KL}(\pi_\theta(y \mid x) \| \pi_{st}(y \mid x)) - \log \frac{\pi_\theta(y \mid x)}{\pi_{st}(y \mid x)} \right) \right) & \text{if } y = y_{\text{u}}
\end{cases}
$$

We set $\lambda_W$ and $\lambda_U$ introduced by \citep{ethayarajh2024kto} to $1$ to weigh chosen and rejected completions equally, and omit them in equation \ref{eq:kto_loss}.

\subsection{Preference dataset}
\begin{figure}
  \includegraphics[width=0.9\linewidth]{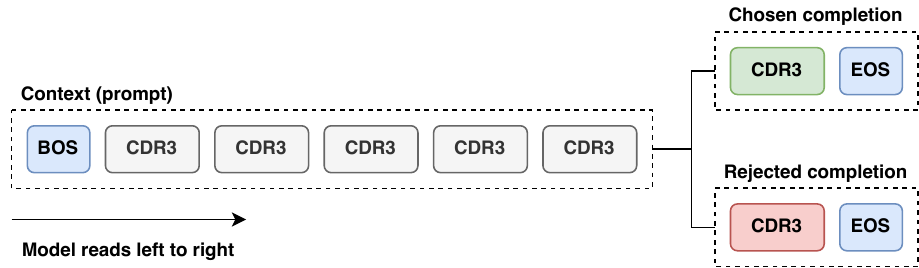}
  \centering
  \caption{Encoding of a pair of chosen and rejected completion for a context prompt. Context- and chosen CDR3s are sampled from good performers for a specific target, rejected CDR3s from poor performers for the same target. We produce up to $n=10$ different pairs for each good CDR3.}
  \label{fig:encoding}
\end{figure}

We construct a preference dataset $\mathcal{D}_p$ based on scored candidates across a number of different target antigens.
We group all candidate sequences by CDR3 (retaining the maximum performing variant of other CDRs if relevant) to form the set of CDR3s $S$ per target antigen, and their associated scores $P$, and split them into good performers (score larger than $t_c$) and poor performers (score less than $t_r$). For every good performer, we construct up to $n=10$ preference pairs by randomly sampling a poor performer and concatenating $k=5$ randomly sampled good performers as context (prompt). Intuitively, these context CDR3s softly encode the target antigen and should allow the model to tailor its conditional distribution to the target during fine-tuning, potentially enabling generalisation to new targets that were not present in the training set (even though this is not the focus of this work).

The difficulty of the task can be controlled via the threshold parameters $t_c$ and $t_r$. Using this procedure and conservative thresholds we construct a total of $70k$ pairs with $3k$ pairs held out as validation set with a fully disjoint set of context and candidate CDR3s. Candidate sequences are restricted to CDR3s and have an average length between $10$ and $11$ amino acids. An example preference pair is illustrated in Figure \ref{fig:encoding}.

\subsection{Fine-tuning \emph{ProGen2} based on preference data}

Throughout this work we rely on \emph{ProGen2} \citep{nijkamp2023progen2} as the pretrained model due to the publicly available implementation and wide choice of pretrained weights, ranging from $151$M to $6.4$B parameters. Pretrained on approximately $280$M protein sequences, ProGen2 serves as a robust baseline for fine-tuning. We do not apply traditional supervised fine-tuning prior to DPO-based training, as we did not observe a tangible benefit in preliminary experiments.

Our implementation is based on Huggingface's DPO trainer, adapting configurations based on model size and hardware constraints. We explore small ($151$M), medium ($764$M) and large ($2.7$B) model variants. Experiments are conducted on Google's Cloud Platform, using $8$ NVIDIA A100 GPUs (80GB) per machine for large models, $4$ for medium models; and a single NVIDIA T4 for small models.

We explore learning rates from \(10^{-4}\) to \(10^{-8}\) and a linear learning rate decay. We use per-device batch sizes of 16 for large models, 32 for medium models, 32 for small models, and gradient accumulation to achieve larger effective batch-sizes. We tune \(\beta\), which controls deviation from the base model, and find \(\beta = 0.1\) works well across most experiments. We train models for approximately 10 epochs, which we found sufficient to ensure convergence, and retain the checkpoint with the lowest validation loss for further experiments.

\begin{figure}
  \includegraphics[width=\linewidth]{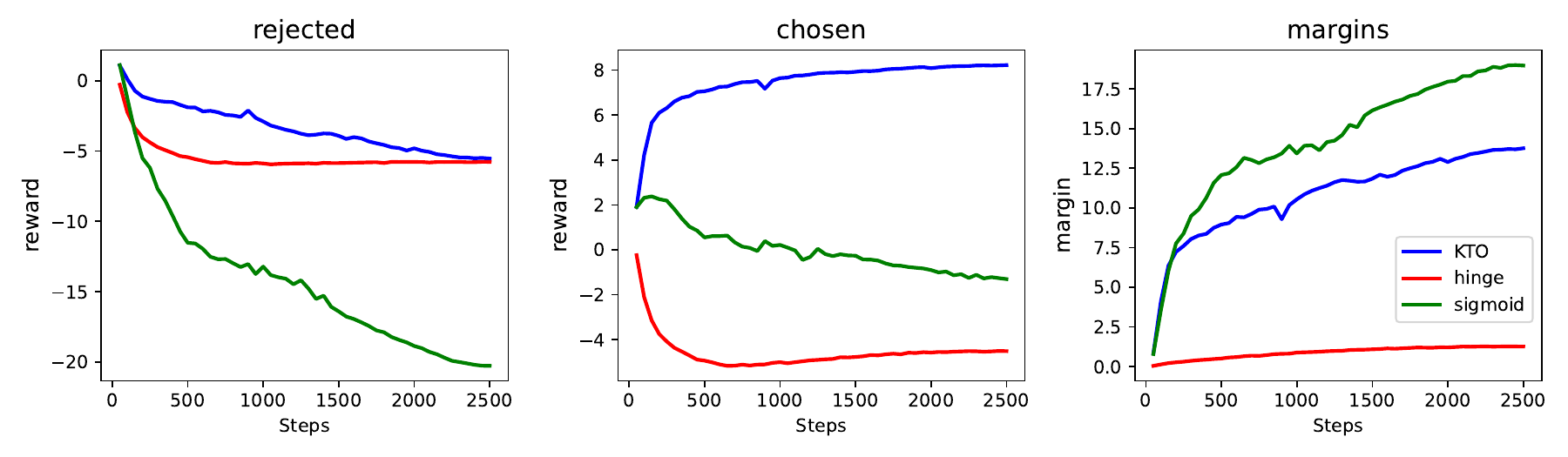}
  \centering
  \vspace*{-1em}
  \caption{Comparison between rewards, the mean difference between the log probabilities of $\pi_\theta$ and  $\pi_\text{st}$ for rejected and chosen completions, and margins between them, for models trained using the three analysed loss functions (best performing hyperparameters were chosen for each loss function).
  Models trained using the original sigmoid-based loss penalise the rejected completions heavily, and tend to become overly confident in their decisions, making them more susceptible to overfitting and numerical instability. They could perhaps benefit from more sophisticated regularisation mechanisms. 
  Models trained with hinge loss maintain very small margins, which may explain their tendency to produce trivial completions, despite promising validation loss and accuracies. It is possible that in the future iterations, after construction of datasets with harder examples \citep{robinson2021contrastive}, models using this loss have some potential to yield useful results due to their rapid convergence and stability.
  KTO's behaviour appears to be best aligned with our desired use-case of hit maturation, as it increases the likelihood of the chosen completion instead of penalising the rejected completion, all while achieving high margins.
  }
  \label{fig:rewards}
\end{figure}

\subsection{Model selection}
We found the medium-sized variant of \emph{Progen2} to be the best trade-off between performance and resource efficiency, where large models did not show improved performance while requiring significantly more compute.

Overall, various combinations of hyper-parameters showed promising training metrics, where we observed validation accuracy beyond 75\%. However, manual inspection revealed that some fine-tuned models produced trivial or collapsed CDR3 sequences when greedily sampling from the model conditioned on a suitable context, which was not reflected in the training metrics. Notably, models trained with hinge loss often overfitted quickly and performed well on the preference dataset, yet generated trivial outputs. We speculate that models trained with KTO loss prioritise the utility of the generated sequences over simply maximising the log-likelihood of preferences, as argued by \citep{ethayarajh2024kto}. This is illustrated in Figure \ref{fig:rewards}. 

The remaining evaluation in this paper is thus focused on a medium-sized model ($764$M parameters) trained with the KTO loss. Training and validation metrics for various learning rates are illustrated in Figure \ref{fig:kto_metrics} in the appendix.

\section{Evaluation}

We aim to evaluate whether a fine-tuned model can be successfully applied to hit maturation, where we explore the design space surrounding a candidate CAR discovered in the pipeline outlined above. Overall we pursue two questions: i) Can we exceed the performance of the candidate CAR by exploring single or double mutants, and ii) is model loss correlated to the performance of CARs in a way that would enable guided exploration of the CAR design space around selected candidates?

We conduct assays in 96-well plates, where each well contains cells expressing a single synthesised CAR with up to two amino acid mutations. This enables us to assess the immune response (activation) individually for each CAR, replicating the best-practice (low throughput) evaluation. We conduct three experiments:

\vspace{-0.5em}
\paragraph{Greedy generation} for two manually selected candidate CARs that showed promising performance in previous experiments for a target included in the training set, we construct a suitable context of $5$ well performing CDR3s, and generate candidates greedily left-to-right, following suggested top $3$ substitutions for each position by the model as outlined in algorithm \ref{algo:greedy}. Based on likelihood (model loss\footnote{We rely on cross-entropy for next-token prediction as a surrogate for likelihood}), we pick the top $15$ single mutants, and the top $15$ double mutants to evaluate on a 96-well plate. Two of the selected mutants were in the training-set, and have been removed from further analysis below.

\vspace{-0.5em}
\paragraph{Exhaustive search} For the same two candidates, we score all possible single and double mutants using the model across all possible permutations of $5$ context CDR3s to get the average likelihood, and pick the top scoring 45 mutants for each candidate to evaluate in a 96-well plate. Three of the selected mutants were in the training-set, and have been removed from further analysis below.

\vspace{-0.5em}
\paragraph{Few shot maturation} To replicate the desired use-case for hit maturation we select $10$ additional candidate CARs (of which $7$ were in the training set). For each CAR, we do an exhaustive search for single and double mutants as outlined above, and evaluate the top $8$ mutants along with each candidate CAR on a 96-well plate. None of the selected mutants were part of the training set.

Note that exhaustive search of all single and double mutants and scoring them across all $120$ context permutations is computationally significantly more expensive than the alternative greedy approach. Intuitively, exhaustive search may find candidates with higher likelihood than the greedy approach, as each position is only conditioned on whatever is "left" of the position, even if a different substitution would significantly increase the likelihood of the remaining positions to the "right".

\subsection{Results}
\begin{figure}
  \includegraphics[width=\linewidth]{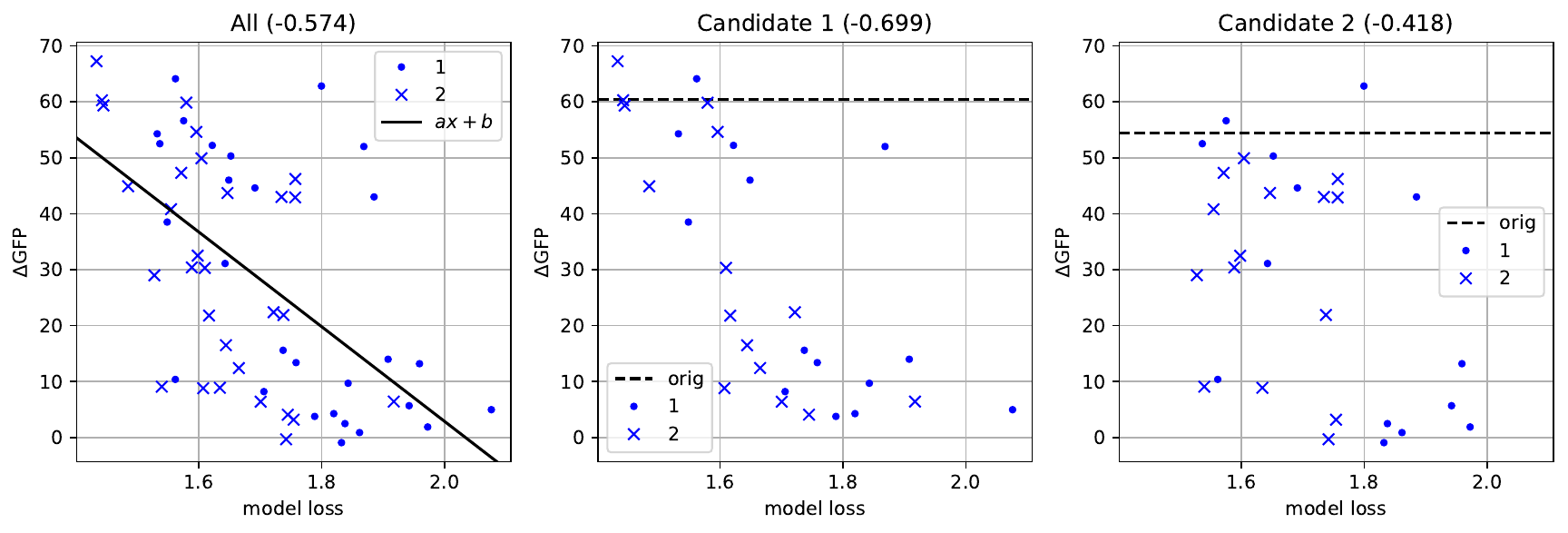}
  \centering
  \vspace*{-1em}
  \caption{Model loss vs activation for greedily generated candidates. All plots show model loss averaged over all possible context permutations, plotted against activation measured as $\Delta$GFP. Dots indicate candidates with a single mutation, crosses two mutations, black lines indicate baseline performance of each candidate. Overall we see strong correlation between averaged model loss and activation.\vspace{-1em}}
  \label{fig:correlation}
\end{figure}

\begin{figure}
  \includegraphics[width=\linewidth]{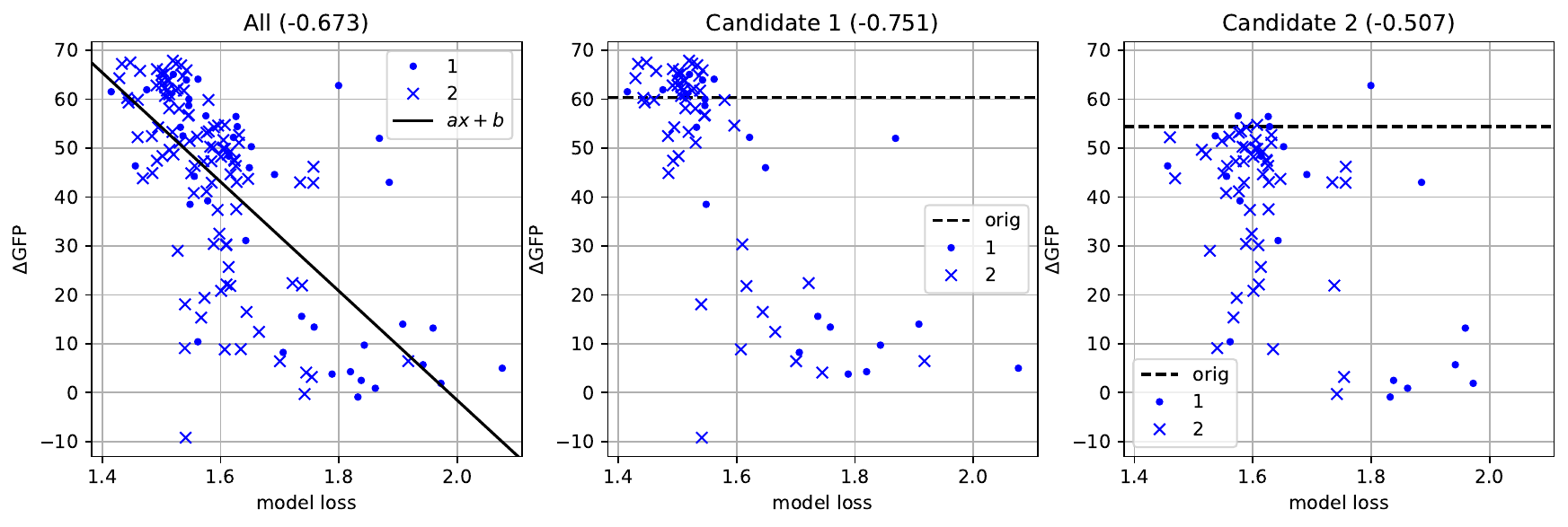}
  \centering
  \vspace*{-1em}
  \caption{Model loss vs activation for greedily generated candidates and candidates from exhaustive search. All plots show model loss averaged over all possible context permutations, plotted against activation measured as $\Delta$GFP. Dots indicate candidates with a single mutation, crosses two mutations, black lines indicate baseline performance of each candidate.\vspace{-1em}}
  \label{fig:correlation_all_combined}
\end{figure}

Figure \ref{fig:correlation} illustrates all single and double mutants discovered using the greedy approach along with a performance score of T cell reporter activation ($\Delta$GFP). Overall, we observe a strong Pearson correlation between average model loss and $\Delta$GFP ($-0.574$). This correlation is higher for Candidate 1 ($-0.699$), but in all cases correlation is significant at $p<0.05$. We also assess the correlation of model loss to $\Delta$GFP of the pretrained model before fine-tuning (see Figure \ref{fig:basemodel} in the Appendix), and find that fine-tuning significantly improves correlation for both candidate 1 ($-0.699$ vs.\ $0.218$) and candidate 2 ($-0.418$ vs.\ $0.072$).

Figure \ref{fig:correlation_all_combined} shows the performance for mutants discovered greedily along with those discovered via exhaustive search. To account for systematic bias, we normalise $\Delta$GFP on the second plate using the performance of each reference CAR as a guide. We see, perhaps surprisingly, that the majority of top candidates found via exhaustive search show promising performance and cluster tightly around the performance of each baseline. In particular, we discover many single and double mutants of candidate 1 that significantly outperform the baseline. Both greedy as well as exhaustive search yield better correlation and more favourable performance for candidate 1, even though both candidates were in the training set (see figure \ref{fig:correlation} and \ref{fig:correlation_all_combined}).

\begin{figure}
  \includegraphics[width=\linewidth]{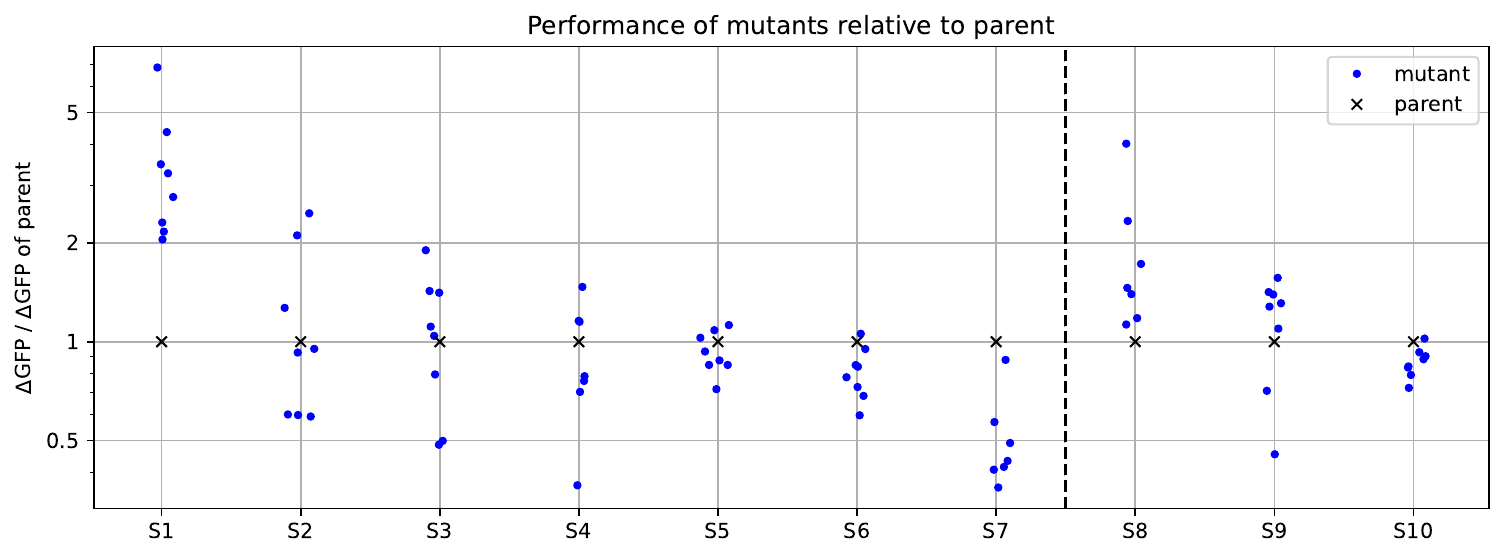}
  \centering
  \vspace*{-1em}
  \caption{For 10 selected sequences, 8 mutants are evaluated in a 96-well plate. The parents in S1-S7 are from the training set, and S8-S10 are from the validation set. None of the mutants occur in the preference dataset.}
\label{fig:p008}
\end{figure}

Figure \ref{fig:p008} shows the performance of mutants relative to their parent for seven candidates from the training set (S1-S7) and three candidates from the validation set (S8-S10). In the majority of cases we discover at least a few mutants that outperform their parent, and in three cases mutants that show more than double the performance -- a promising result given the size of the search space ($10^4$ - $10^5$). Performance further appears to generalise to the validation set (S8-S10), but we do not yet have sufficient experimental data to determine statistical significance.

Further work is necessary to explore the factors that influence the performance of the approach, to determine if the source of the discrepancy between candidate 1 and candidate 2 is biological (i.e.\ structural), or whether it may be an artefact from the construction of the preference dataset. While these preliminary results are difficult to interpret due to low number of candidates, we can draw two conclusions: i) Overall model loss seems to be well correlated to candidate performance, which supports our hypothesis that such models can guide exploration of the design space surrounding a candidate CAR, and ii) to maximise the chances of discovering high-performing mutants, it is crucial to conduct an exhaustive search, even if this comes at higher computational cost. Recent advances in sampling from PLMs \citep{hayes2024simulating}, may address this limitation, which we leave for future work to explore. 

\subsection{Discussion}

In this work we explore Protein Language Models (PLMs) for hit maturation in cell therapy, and demonstrate how data from a high-throughput experimental platform can be harnessed to formulate a preference task. Even with relatively simple experiments, using well-established methods and pre-trained publicly available models, we obtain promising results -- highlighting the significant potential of PLMs in cell therapy, which could generalise to other therapeutic modalities.

As the presented results are preliminary there remains much to explore, including leveraging multi-dimensional single-cell data to capture more comprehensive information on gene expression profiles, which would potentially alleviate the lack of precision of high-throughput testing. Furthermore, our approach is agnostic to spatial protein structure, where Graph Learning techniques \citep{hetzel2021graph, jiang2024gte} could lead to improved performance and better understanding of which candidates have the most potential for maturation.

\begin{algorithm}[t]
    \caption{Greedy diversification}
    \small
    \begin{algorithmic}
    \Function{$\mathrm{GREEDYGEN}$}{$L, R, k, \mathrm{topk}$}
        \State $D = [\mathrm{concat}(L, R)]$
        \If{$k == 0$ or $|R| < 2$}
            \State return $D$
        \EndIf
        \State $\text{S} = [ \argmaxtopk\!\left( \mathrm{logits}(\mathrm{concat}(L, R))_p \right) \forall p \in [1 \dots (|R|\!-\!1)]]$ \Comment{get top k subst.\ per position in $R$}
        \For{$p \in [1 \dots (|R|\!-\!1)]$}
            \For{$s \in S_p$}
                \If{$\mathrm{valid}(s)$}
                    \State $\mathrm{D} = D \cup \mathrm{GREEDYGEN}(\mathrm{concat}(L, R_{\mathrm{:}p}, s), R_{(p+1)\mathrm{:}}, k-1, \mathrm{topk} )$
                \EndIf
            \EndFor
        \EndFor
        \State $\mathrm{return}(D)$
    \EndFunction
    \end{algorithmic}
    \label{algo:greedy}
\end{algorithm}

\bibliography{main}

\appendix

\section{Appendix}
\begin{figure}[h]
  Training
  \includegraphics[width=\linewidth]{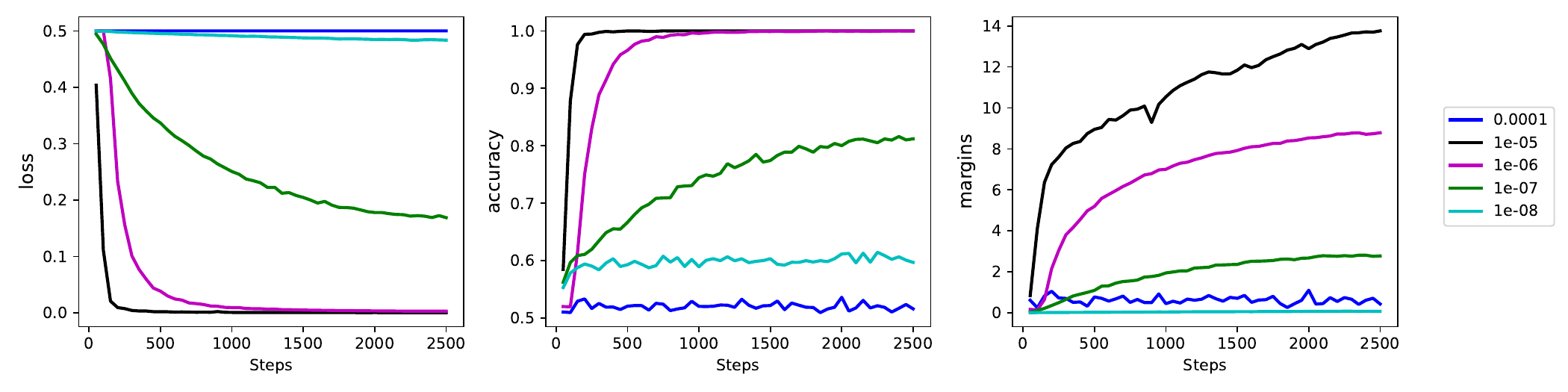}
  Validation
  \includegraphics[width=\linewidth]{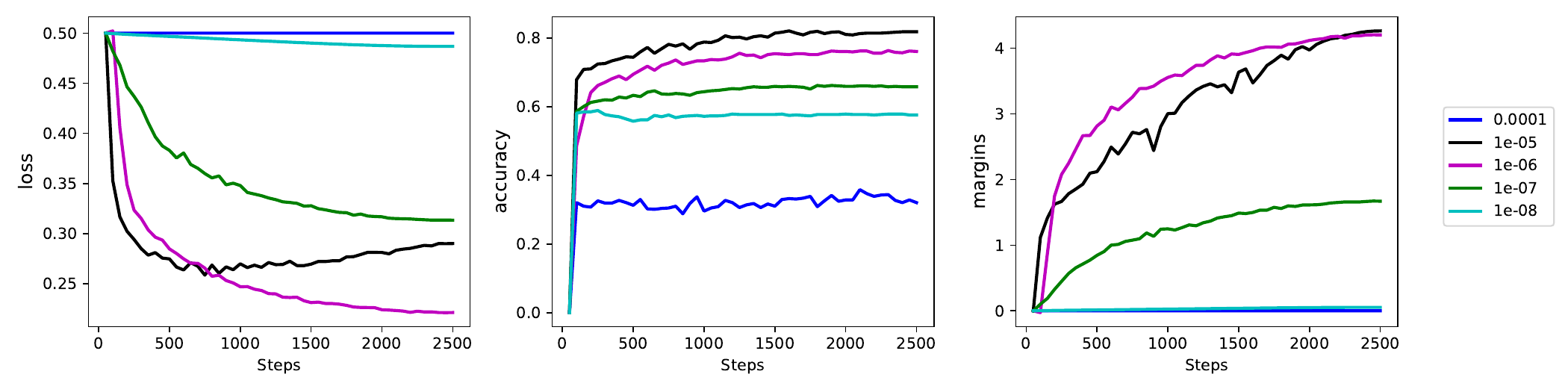}
  \centering
  \vspace*{-1em}
  \caption{Training (top row) and validation metrics (bottom row) for various settings of learning rate in a \emph{progen2-medium} trained with KTO. Loss remains high for both the smallest and largest learning rates. Based on these results and further qualitative evaluation (see text for details), we selected $10^{-5}$ for further experimentation in this work.}
\label{fig:kto_metrics}
\end{figure}

\begin{figure}[h]
  \includegraphics[width=0.9\linewidth]{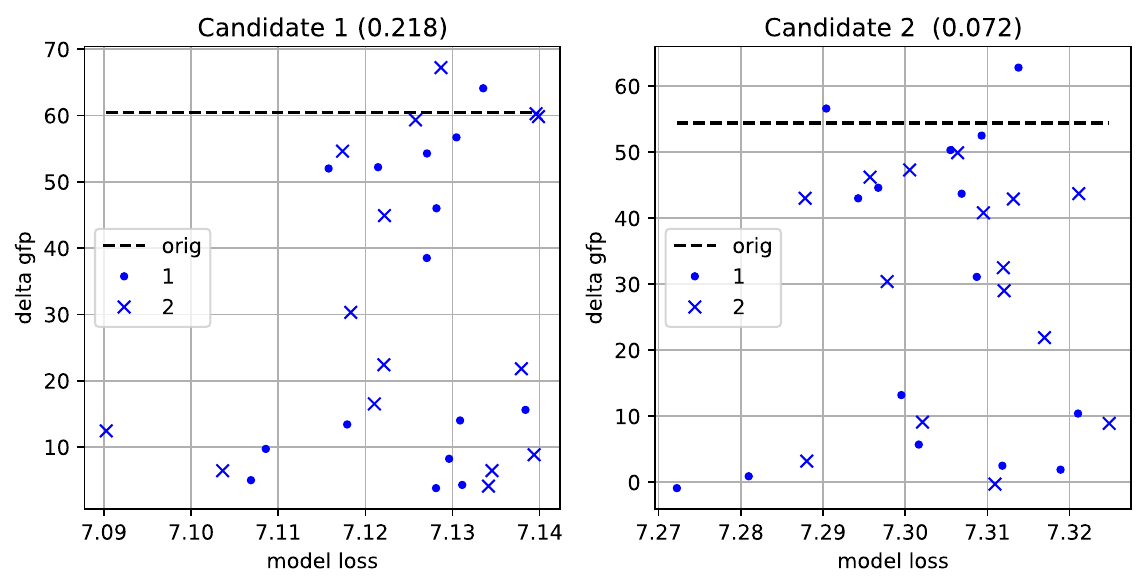}
  \centering
  \vspace*{-1em}
  \caption{Model loss of \emph{progen2-medium} before fine-tuning vs activation for greedily generated candidates. All plots show model loss averaged over all possible context permutations, plotted against activation measured as $\Delta$GFP. Dots indicate candidates with a single mutation, crosses two mutations, black lines indicate baseline performance of each candidate. As expected, performance before fine-tuning is poor (insignificant Pearson correlation of $0.218$ and $0.072$), where some of the worst performing mutants obtain the lowest loss. Without fine-tuning, this model would not be helpful when exploring the design space of CARs, motivating the need for preference-based fine tuning of PLMs.\vspace{-1em}}
\label{fig:basemodel}
\end{figure}

\end{document}